\documentclass[letterpaper, conference, 10pt]{IEEEconf}
\IEEEoverridecommandlockouts

\usepackage[tracking=true]{microtype}
\usepackage{authblk}
\usepackage{amsmath} 
\usepackage{amssymb}  
\usepackage{amsfonts}
\usepackage{graphicx}
\usepackage[linesnumbered,ruled,vlined]{algorithm2e} 
\usepackage{algorithmic}
\usepackage{epsfig} 
\usepackage{times} 
\usepackage{xspace}
\usepackage{url}
\usepackage{comment}
\usepackage[dvipsnames]{xcolor}
\usepackage[font={footnotesize}]{caption}
\usepackage{booktabs} 
\usepackage{tikz}
\usetikzlibrary{positioning,shapes.geometric,arrows,fit,shapes.symbols,svg.path,tikzmark,calc}
\usepackage{textcomp}
\usepackage{listings}
\usepackage{subcaption}
\pgfdeclarelayer{background}
\pgfdeclarelayer{foreground}
\pgfsetlayers{background,main,foreground}
\usepackage{adjustbox}
\usepackage{array}
\usepackage{multirow}
\usepackage{dsfont}

\usepackage{siunitx}
\usepackage{mathtools}

\usepackage{physics}

\let\labelindent\relax 

\usepackage[inline]{enumitem} 

\usepackage{amsthm}

\usepackage[colorlinks,
            linkcolor=blue,
            citecolor=MidnightBlue,
            urlcolor=blue]{hyperref}
            
\usepackage[backend=biber,
            hyperref=true,
            url=false,
            isbn=false,
            doi=false,
            backref=false,
            style=ieee,
            natbib=true,
            mincitenames=1,
            maxcitenames=1,
            citestyle=numeric-comp,
            sorting=none,
            block=none]{biblatex}

\newcommand{\conf}{C_t}

\newtheorem{Tho}{Theorem}

\newtheorem{Def}{Definition}

\newcommand{\samp}{LICR-CS }

\SetKwInput{KwInput}{Input}                
\SetKwInput{KwOutput}{Output}     
            
\renewcommand{\bibfont}{\small}
\newcommand{\algname}{LAVQA\xspace}
\newcommand{\mapname}{LICOM\xspace}

\usepackage{pifont}

\usepackage{float}
\title{\LARGE \bf
\algname: A Latency-Aware Visual Question Answering Framework \\ for Shared Autonomy in Self-Driving Vehicles
}
\author{Shuangyu Xie$^{1,*}$,
Kaiyuan Chen$^{1,*}$,
Wenjing Chen$^{4}$,
Chengyuan Qian$^{4}$,
Christian Juette$^{2}$,\\
Liu Ren$^{2}$,
Dezhen Song$^{3}$,
Ken Goldberg$^{1,5}$\\
\footnotesize
$^{1}$University of California, Berkeley \quad
$^{2}$Bosch Research \quad
$^{3}$MBZUAI \quad
$^{4}$Texas A\&M University \quad
$^{5}$IEOR, UC Berkeley
$*$ indicate equal contribution
}

\begin{document}

\twocolumn[{%
\renewcommand\twocolumn[1][]{#1}%
\maketitle
\vspace{-20pt}
\begin{center}
    \centering
    \captionsetup{type=figure}
    \includegraphics[width=6.5in]{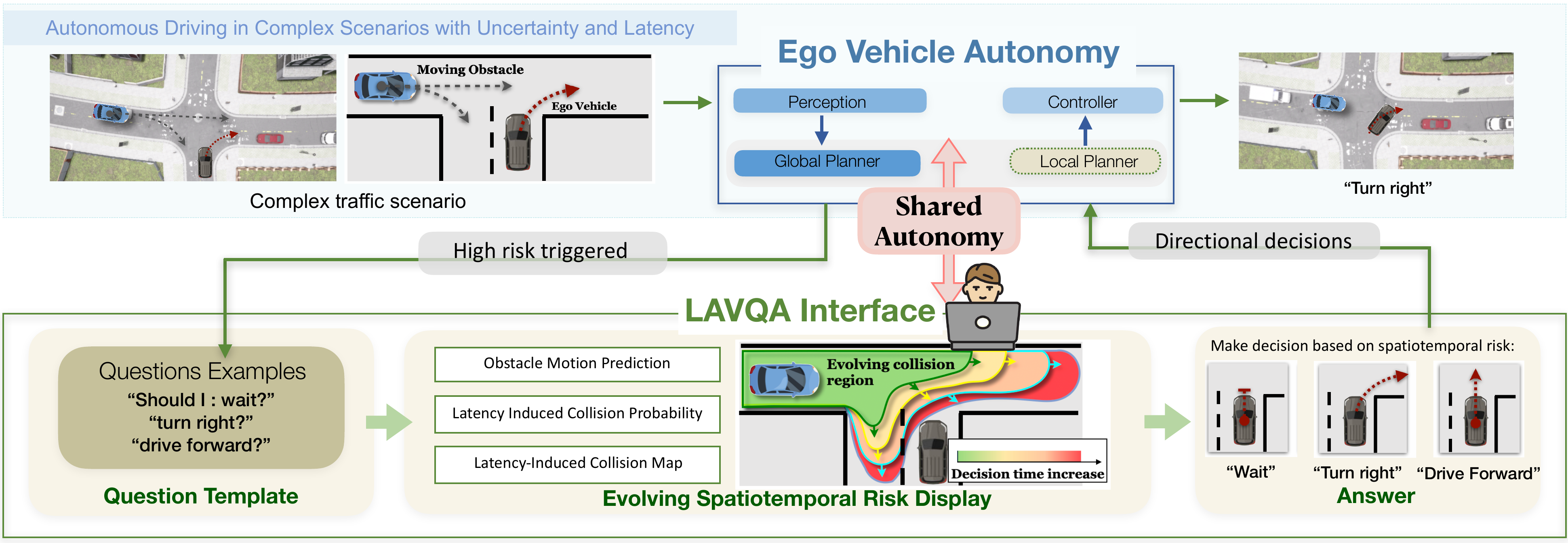}
    \captionof{figure}{\textbf{\algname} is a latency-aware shared autonomy framework that the remote operator and autonomous vehicle collaboratively make safe and context-appropriate decisions. It uses visual question answering and visually augments how collision risk evolves over decision latency to assist in vehicle navigation.}
    \label{fig:scene-title}
\end{center}%
}]


\begin{abstract}
When uncertainty is high, self-driving vehicles may halt for safety and benefit from the access to remote human operators who can provide high-level guidance. 
This paradigm, known as \textit{shared autonomy}, enables autonomous vehicle and remote human operators to jointly formulate appropriate responses.
To address critical decision timing with variable latency due to wireless network delays and human response time, 
we present \algname, a latency-aware shared autonomy framework that integrates Visual Question Answering (VQA) and spatiotemporal risk visualization.
\algname augments visual queries 
with Latency-Induced COllision Map (\mapname), a dynamically evolving map that represents both temporal latency and spatial uncertainty. 
It enables remote operator to observe as the vehicle safety regions vary over time in the presence of dynamic obstacles and delayed responses.
Closed-loop simulations in CARLA, the de-facto standard for autonomous vehicle simulator, suggest that that \algname can reduce collision rates by over 8x compared to latency-agnostic baselines.

\end{abstract}

\section{Introduction}

Autonomous Vehicles (AVs) are increasingly deployed in complex and dynamic environments
~\cite{van2018autonomous, prakash2021multi,claussmann2019review, schwarting2018planning, nayakanti2023wayformer,kuutti2020survey}. 
In some complicated or ambiguous cases, such as construction zone or passing stalled vehicles,  vehicle on-board planning algorithms may still face high uncertainty.  
Many AV systems, such as Waymo~\cite{waymo_fleet_response}, halt to ensure safety and request assistance from a remote human operator. 
They transmit the visual scene context and solicit human guidance through high-level intent or explicit future waypoints~\cite{waymo_fleet_response}.
This paradigm, known as \textit{shared autonomy}, enables vehicles to handle out-of-distribution scenarios that exceed the capabilities of current algorithms ~\cite{javdani2015shared, jeon2020shared, nikolaidis2017human}.
However, these human-in-the-loop decisions are subject to variable latencies due to both human cognition and network transmission. 
The latency reduces the time window and thus shrinks the spatial region available for AV safe maneuvering in the presence of dynamic obstacles. This raises a critical design question: 
How can we facilitate decision-making that accounts for both \textit{spatial} safety and \textit{temporal} delay?


As an example in Figure~\ref{fig:scene-title}, consider an AV waiting at an unregulated intersection with dense cross-traffic and pedestrians. The vehicle detects surrounding dynamic obstacles but remains uncertain whether it is safe to proceed. To resolve this, the AV can request  assistance by querying remote operator with {Visual Question Answering (VQA)},  such as “Is the gap sufficient to turn?”. The operator responds with a high level intent  (e.g., “proceed” or “yield”) or an explicit waypoint indicating whether to commit to the turn or wait. However, the response may arrive too late after the latest feasible moment when braking or yielding can still prevent a collision, leading to a dangerous maneuver.
This motivates the need to model how delayed response can affect AV safety. 
We consider the following conditions: 
(a) the ego vehicle may query remote operator with a Visual Question based on the current scene context;
(b) the remote operator’s response—whether an answer or a waypoint selection—is subject to variable decision latency from processing time and network transmission;
(c) the environment contains dynamic obstacles (e.g., vehicles, pedestrians) with uncertain motion, which evolve over time during the decision delay. 

This setup is challenging for conventional shared autonomy frameworks in AV. Existing collision assumes decision latency to be either negligible or constant. Similarly,
conventional VQA frameworks, such as EMMA~\cite{hwang2024emma}, evaluate performance using spatial displacement or task success metrics,  However, variable network delays impact the decisions, see Table \ref{tab:latency_only_prompts}. 

We propose \algname, a latency-aware shared autonomy framework for AVs. 
To facilitate the operator to provide decisions that are both spatially and temporally safe, 
\algname introduces Latency-Induced Collision Map (\mapname), a dynamically evolving spatiotemporal map that models both temporal latency and spatial uncertainty. 
By augmenting visual questions with 
\mapname, \algname illustrates how vehicle safety regions vary over time in the presence of dynamic obstacles and decision delay.
 

We systematically study how decision delays impact spatial safety using closed-loop simulation in CARLA~\cite{dosovitskiy2017carla}, a photorealistic and industrial-grade autonomous driving simulator. 
Experimental results suggest that under a simulated human decision latency, \algname reduces collision rates by up to 8.98x compared to a latency-agnostic baseline.

This paper makes three contributions: 
\begin{enumerate*}
    \item \algname, a shared autonomy framework for Autonomous Vehicles that integrates remote operator response latency;
    \item \mapname, A latency-aware safe region visualization that constructs temporally growing collision zones;
    \item Evaluation data suggesting \algname reduces collision rate by 8.9x.
\end{enumerate*}

\section{Related Work} \label{Sec:real-time}




\textbf{Autonomous Driving Safety}
Autonomous vehicles has been increasingly deployed with advanced perception\cite{kirillov2023segment}, planning\cite{highway_review,schwarting2018planning}, control\cite{bonzanini2021perception}, and end-to-end learning frameworks \cite{chen2024end} to predict dynamic environment evolution\cite{gao2020vectornetencodinghdmaps,nayakanti2023wayformer}, compute collision probabilities\cite{Jasour-RSS-21}, and execute avoidance maneuvers.
However, complex road scenarios may exceed the safety threshold of current algorithms, resulting in out-of-distribution cases \cite{bogdoll2022anomaly} where the likelihood of unsafe or dangerous outcomes increases \cite{antonante23rss}.
In such high-uncertainty situations, human supervision is often required to resolve the issue and replan accordingly.

\textbf{Visual Augmentation in Shared Autonomy}
Shared autonomy is a paradigm that connects remote human operators to autonomous systems, such as robots or vehicles, and enables the operator to issue actions that are executed by the system~\cite{javdani2015shared, jeon2020shared, nikolaidis2017human}. 
Human operators primarily rely on visual information to make decisions, but accurately interpreting spatial layouts and dynamics in unfamiliar environments can be challenging. 
Visual augmentation has been shown to enhance decision-making accuracy and task success rates. For example, Brizzi et al.~\cite{brizzi2017effects} demonstrated that augmented reality (AR) overlays significantly improved the performance of remote industrial assembly tasks. 
In the context of driving, Graf et al.~\cite{graf2020predictive} proposed predictive visual overlays to assist operators in teleoperated vehicle navigation.
Although humans are capable of making informed decisions, they are subject to cognitive and perceptual latency~\cite{schitz2021shared}.
In highly dynamic and time-critical scenarios such as autonomous driving, visual augementation  can serve not only to improve spatial understanding but also to reflect how risk evolves with operator delay. 
While prior work (such as lunar rover teleoperation~\cite{litaker2025lunar}) has considered latency effects, it often neglects the safety-critical implications of delayed decisions. 
In contrast, \algname introduces a visual augmentation framework that explicitly models and communicates the spatiotemporal evolution of collision risk.

\textbf{VQA in Autonomous Driving~}  
VQA provide interactive paradigm through asking any type of question about an image for which the human or system should return the correct answer~\cite{antol2015vqa}. VQA has been widely adopted across various domain such as image captioning \cite{zhou2020unified}, video understanding \cite{jang2019video}, and robotic task planning for  manipulation~\cite{robo2vlm, sermanet2024robovqa} and navigation \cite{buoso2025select2plan}. Recently, it is often designed as a benchmarking format to evaluating the visual reasoning capabilities of Visual Language Model \cite{xing2025can, goyal2017making, lee2022what}.
For self-driving scenarios, VQA is an emerging format that connects scene observations from sensory data with natural language to support driving-relevant question answering~\cite{deruyttere2019talk2car,hwang2024emma,xing2025openemma}. In driving tasks, the questions typically follow a predefined template, and the answers take a closed-form structure that maps directly to control commands or high-level executable decisions for the vehicle. \algname follows this same paradigm, serving as a decision-making interface for self-driving.

\begin{table}[t]
\centering
\resizebox{\columnwidth}{!}{%
\begin{tabular}{p{0.58\linewidth}cc}
\toprule
\textbf{Query} & \textbf{Answer at $t$} & \textbf{Answer at $t+\tau$} \\
\midrule
Is it safe to merge in front of the blue car?                         & \textit{Proceed}  & \textit{Yield} \\
Is the current left-turn gap sufficient to cross?                     & \textit{Go}       & \textit{Wait} \\
Can I turn right before cross-traffic?                 & \textit{Turn}     & \textit{Yield} \\
Can I enter the intersection before the pedestrian reaches the crosswalk? & \textit{Proceed}  & \textit{Stop} \\
Is it safe to pass the parked truck now?                              & \textit{Pass}     & \textit{Hold} \\
On-ramp gap selection: should I merge now?                            & \textit{Merge}    & \textit{Hold} \\
Overtake the cyclist before the lane narrows?                         & \textit{Yes}      & \textit{No} \\
Lane closure ahead. Should I use left lane?                                 & \textit{Yes}      & \textit{Wait} \\
\bottomrule
\end{tabular}%
}
\caption{\textbf{VQA Examples that the decision latency could induce different decisions for safety} Each question is common for autonomous driving that the recommended action at time $t$ can differ from executing the same decision after a delay $\tau$.}
\label{tab:latency_only_prompts}
\end{table}

\section{Problem formulation}
We formulate the latency visualization problem that augment shared autonomy for autonomous driving and deal with dynamic obstacles. We start with assumption:  


\subsection{Assumption}
\begin{itemize}
    \item[a.1] For ground vehicles, the environment is represented in a bird’s-eye view (BEV). The configuration space of the ego vehicle and surrounding obstacles is restricted to the 2D plane, where each vehicle and obstacle is in $SE(2)$.
    \item[a.2] An ego vehicle autonomy stack generates a reference trajectory and decision template. 
    \item[a.3] Each obstacle follows a stochastic motion model that captures uncertainty in its future trajectory~\cite{Patil_icra2012}. 
\end{itemize}

\subsection{Latency Definition} \label{sec:latency}
We define \textit{decision latency} $\tau$ as the total delay between the moment an AV issues a query and the moment the corresponding action is executed. This latency consists of two components:  
(a) Human response latency: the time required for a remote operator to perceive the visual context, interpret the query, and provide a decision (e.g., intent selection or waypoint specification).  
(b) Network transmission latency: the time required to transmit visual query and contextual information from the vehicle to the operator and to return the operator’s response back to the vehicle.  The overall latency is therefore sum of the both latencies. 

\subsection{Shared Autonomy}
In shared autonomy, the remote human operator observes the  scene at time $t$ and makes a  decision  $a_t \in \mathcal{A}$ to select a directional command, such as lane changes or turns. 
This decision reflects a \textit{best-effort} judgment that appears both safe and conducive to progress.
The decision will be executed after a latency $\tau$, taking effect at the decision time $t_d = t + \tau$.

For ego vehicle and dynamic obstacle in the scene, we define a probability measure that reflect safety. 
Let $\mathcal{E}_{t_d}(a_t)$ denote the predicted occupancy region of the ego vehicle after executing $a_t$ at time $t_d$, and let $\mathcal{O}_{t_d}$ denote the predicted occupancy region of surrounding obstacles at that time.
\begin{Def}[\textbf{Decision Safety Measure}] We define probability that no collision occurs at the time of execution:
\begin{equation}
\mathcal{S}(a_t, \tau) := \mathds{P} \left( \mathcal{E}_{t_d}(a_t) \cap \mathcal{O}_{t_d} = \emptyset \right).
\end{equation}
This measure satisfies $\mathcal{S}(a_t, \tau) \in [0, 1]$, where higher values indicate greater safety.
\end{Def}

Note that due to the uncertainty in obstacle motion intention, the absolute safe cannot be guaranteed. We define a safety threshold $\lambda$ and the action is considered as safe if it satisfy the following condition:
\begin{Def}[\textbf{$(\lambda,\tau)$-Safety}]\label{safety}
 A decision $a$ is considered $(\lambda,\tau)$-safe if $\mathcal{S}(a, \tau) > 1 - \lambda$.
\end{Def}


Achieving $(\lambda,\tau)$-safety is challenging, as human operators must make decisions based on an intuitive understanding of the current scene, without precise or quantitative tools to estimate collision probability. This difficulty is further compounded by the need to anticipate how the scene will evolve during the decision latency and how their chosen action will interact with that evolution. 

We study the problem, given the current scene, how to provide a $(\lambda,\tau)$-safety measure and help human make best-effort decision in the presence of latency $\tau$.

\section{\algname}

We propose \algname, a latency-aware shared autonomy framework for AV that assists human operators in making spatially and temporally safe decisions.
\algname uses VQA to enable operators quickly react to dynamic driving scenes and augments VQA with spatiotemporal risk visualizations for $(\lambda,\tau)$-safety. 
When the ego vehicle encounters ambiguous or uncertain situations, it can query a remote human operator for guidance with \algname interface augmented with spatiotemporal risk. 
After the human operator makes the decision, 
\algname forwards the decision to the vehicle on-board local planner and controller for execution.

The overall architecture is shown in Figure \ref{fig:scene-title}. The ego vehicle operates with a standard autonomy stack consisting of perception, global planner, local planner, and low-level controllers. The perception module takes as input raw sensor data (e.g., RGB images) and outputs a semantic and geometric scene representation; the global planner consumes the scene representation and high-level goals to produce a nominal route or trajectory; the local planner refines this trajectory in response to real-time conditions and human input, generating a latency-aware, safety-validated motion plan;
and the low-level controllers convert the motion plan into executable control commands (e.g., throttle, steering) for the vehicle actuators.

\subsection{VQA For AV Shared Autonomy}
\algname uses VQA as a shared autonomy interface to facilitate time-critical and safe decisions between human and autonomous vehicles. 
At time $t$, a Visual Question consists of visual input of a BEV of the scene, and associated textual questions generated in response to planner uncertainty or conflict events. The operator’s response is constrained to a space of directional decisions $a_t$.
For example, when the global planner generates a tentative trajectory and a pedestrian is detected approaching a crosswalk while the vehicle prepares to turn,
\algname captures the current scene state, extracts the relevant geometric and semantic information, and generates a concise, scene-specific question from template. 
The generated question might be ``Can I turn before the pedestrian reaches the crosswalk?’’ with the corresponding decision space being {turn} and {stop}. These responses are parsed by \algname and relayed to the on-board local planner and controller for execution.

\subsection{VQA Triggering Condition and Execution}

The planner continuously monitors motion predictions and risk estimates to determine when operator intervention is required. 
Each VQA instance in \algname is initiated by a triggering condition that the ego vehicle enters a high-uncertainty region or detects an obstacle with increasing collision probability. 
After the operator provides a response, the system parses the answer into a concrete control command. If the response is discrete, such as ``stop’’, the local planner generates a deceleration profile that yields before the obstacle. If the response is continuous, such as a new waypoint, it is integrated into the planner’s trajectory buffer and fused with the existing plan. 
The updated trajectory is then executed using standard low-level controllers, such as Model Predictive Control (MPC) or PID. Before execution, the local planner verifies the feasibility of the operator’s decision by checking whether the action corresponds to a reachable future state, and rejecting commands that are either outdated (such as past waypoints) or physically infeasible under current constraints.

\subsection{Spatiotemporal Risk Visual Augmentation}
To help the operator understand how collision risk evolves under delayed responses, \algname computes a spatiotemporal risk map over a discretized grid on BEV. 
The resulting latency-induced map represents a mapping from a spatial grid in the 2D plane to a measure reflects the safety for the ego vehicle traveling on the grid given the decision latency. The latency induced map is then projected back into the image space and rendered as a visual overlay on the BEV camera view, highlighting regions of high future risk. These augmentations dynamically changes over time to provide intuitive visual feedback on how the set of safe responses changes with increasing response latency.

In Section \ref{sec:method}, we design a instance for the latency risk map using the collision probability as the safety measure, we will detail the formulation and derivation of the risk and risk map for visualization.

\section{Latency Induced Collision Map}
\label{sec:method}
In this section, we present the derivation of the Latency-Induced Collision Map (LICOM), a spatiotemporal risk representation that uses collision probability as the safety measure $\mathcal{S}(a, \tau)$. As illustrated in Fig.~\ref{fig:scene-title}, our approach first predicts obstacle trajectories and then estimates time-varying collision probabilities—referred to as the Latency-Induced Collision Probability (LICP). LICOM visualizes how collision risk evolves with increasing decision latency, enabling human operators to make spatially and temporally informed decisions.


\subsection{Motion Prediction}
 Denoted the current pose for obstacle $i$ as $\mathbf{x}_i(t) = [x,y,\theta]^{\mathsf{T}} \in SE(2)$, where $t$ is the current time. For each dynamic obstacle $i$, we predict its future pose.
To capture the uncertainty in obstacle motion, we assume each obstacle follows a multi-modal Gaussian distribution:  
\[
\mathds{P}(\mathbf{x}_i(t)) = \sum_{j=1}^J \Psi_j \,\mathcal{N}(\mathbf{x}_i(t)\,|\,\mu_{j,t}, \Sigma_{j,t}),
\]  
where $\mu_{j,t}$ and $\Sigma_{j,t}$ denote the mean pose and covariance of mode $j$, and $\Psi_j$ are mixture weights satisfying $\sum_j \Psi_j = 1$.  The state transition model is then given by  
\begin{equation}\label{eq:predict}
\mathds{P}(\mathbf{x}_i(t+\delta t)\mid \mathbf{x}_i(t)) = \sum_{j=1}^J \Psi_j(\mathbf{x}_i(t)) \,\mathcal{N}(\mathbf{x}_i(t+\delta t)\,|\,\mu_{j,t}, \Sigma_{j,t}).
\end{equation}  

This Gaussian mixture model (GMM) captures multiple possible futures for each obstacle. The parameters can be estimated using methods such as the unscented Kalman filter (UKF)~\cite{UKF-human-predict}, Bayesian multi-policy prediction~\cite{galceran2015multipolicy}, or deep learning approaches with structured map encoding~\cite{gao2020vectornetencodinghdmaps} and conformal planning~\cite{pmlr-v211-dixit23a}. This prediction step provides the foundation for evaluating collision risk in dynamic traffic scenarios.  

\subsection{Collision Probability Estimation}
We first define the collision probability.
\begin{Def}[\textbf{Collision Probability}] \label{def:collision-def-general}
Let $\mathcal{E} \subset \mathbb{R}^2$ and $\mathcal{O} \subset \mathbb{R}^2$ denote the regions occupied by the ego vehicle and an obstacle, respectively. At time $t$, define a random variable $C(t)$ indicating a collision event:  
$$C(t) = 
\begin{cases}
1, & \text{if } \mathcal{E} \cap \mathcal{O} \neq \emptyset, \\
0, & \text{otherwise}.
\end{cases}$$ The collision probability is   
\begin{equation}\label{eq:collision_def_set}
\mathds{P}(C(t)=1) = \mathds{P}\big(\mathcal{E} \cap \mathcal{O} \neq \emptyset \big).
\end{equation}
\end{Def}  

\noindent Let $\mathcal{E}(\mathbf{x}_0(t))$ denote the ego vehicle’s occupied region and $\mathcal{O}(\mathbf{x}_i(t))$ the obstacle’s region at time $t$. For a single obstacle, the collision probability is  
\begin{align}\label{eq:col_prob}
\mathds{P}_{c}(t) &= \mathds{P}\big(\mathcal{E}(\mathbf{x}_0(t)) \cap \mathcal{O}(\mathbf{x}_i(t)) \neq \emptyset\big) \\ 
&= \int \mathbb{I}_C(\mathbf{x}_i(t), \mathbf{x}_0(t)) \,\mathds{P}(\mathbf{x}_i(t), \mathbf{x}_0(t)) \,\mathrm{d}\mathbf{x}_i(t), \label{eq:collision-prob}
\end{align}  
where $\mathbb{I}_C$ is the indicator function:  
\[
\mathbb{I}_C(\mathbf{x}_i(t), \mathbf{x}_0(t)) = 
\begin{cases}
1, & \text{if } \mathcal{E}(\mathbf{x}_0(t)) \cap \mathcal{O}(\mathbf{x}_i(t)) \neq \emptyset, \\
0, & \text{otherwise}.
\end{cases}
\]  

In practice, $\mathds{P}_{c}(t)$ can be approximated by evaluating the probability of non-positive displacement between the closest points of the ego vehicle and the obstacle. Efficient convex hull approximations for these computations are described in~\cite{convex_hull_col_prob}.

\subsection{Latency-Induced Collision Probability (LICP)}
Building on the time-indexed motion prediction and collision probability estimation, we now quantify how risk evolves under delayed decision-making. 
We define the LICP as the estimated collision probability at a delayed moment 
conditioned on the current states of the ego-vehicle and the obstacles. This is a derivation of $\mathcal{S}(a,\tau)$. Ego vehicle pose is $x_0(t_d) = f(x_0(t),a)$, where $f$ is a mapping for the state transition function given the human decision $a$.

\newcommand{\licp}[2]{\ensuremath{\widehat{\mathds{P}}_{c}(#1,#2)}}
\begin{Def}[Latency-Induced Collision Probability]\label{def:licr}
The ego-vehicle's pose at time $t_d$ following the current trajectory
is $x_0(t_d)$.
The latency-induced collision probability $\licp{t}{\tau}$ is defined as
\[
\licp{t_d}{x_0(t_d)} =\mathds{E}\big[ \mathds{P}_c(t_d)\mid x_0(t_d), \mathbf{x}_i(t) \big]
\]
Where $ \mathbf{x}_i(t)$ follows the distribution $\mathds{P}(\mathbf{x}_i(t))$.
\end{Def}

LICP entails the  state distribution, the transition model for obstacles and the collision probability calculation. 
To illustrate this, we unfold the formulation in Definition \ref{def:licr}
as follows:
\begin{multline}
\licp{t_d}{x_0(t_d)} =\int \mathds{P}(\mathbf{x}_i(t))\mathrm{d}\mathbf{x}_i(t) \\ \cdot \int \mathbb{I}_C(\mathbf{x}_i(t_d ), \mathbf{x}_0(t_d )) \cdot \mathds{P}(\mathbf{x}_i(t_d ) \mid \mathbf{x}_i(t))\mathrm{d}\mathbf{x}_i(t_d) 
\cdot
\label{eq:lico-integral}
\end{multline}
Where the first integration calculates the collision probability
with the distribution of $\mathbf{x}_i(t_d )$ estimated by Equation \eqref{eq:predict}, while the second integrates over the current state distribution.
 This formulation accounts for the decision latency and incorporates it as a tunable variable. For situation with multiple dynamic obstacle, we only consider the one closest to the ego vehicle. 

\subsection{Visualization}

To spatially represent how collision  evolves under decision latency, we define a {Latency-Induced Collision Map(LICOM)}. This map overlays the 2D workspace with a discretized grid, where each cell encodes the probability of a collision occurring at that spatial location, given a latency duration $\tau$. LICOM generalizes classical occupancy grids by embedding both spatial uncertainty (due to obstacle motion) and temporal uncertainty (due to delayed execution). 

\begin{Def}[LICOM] \label{def:lico-map}
Let the 2D drivable space be discretized into a finite grid $\mathcal{G} = \{g_1, g_2, \ldots, g_M\}$, where each cell $g_m \in \mathbb{R}^2$ corresponds to a spatial region (e.g., square or rectangle) on the map. 
\[
\text{LICOM} : \mathcal{G}\times R_{>0} \rightarrow [0, 1]
\]
defined for each grid cell $g \in \mathcal{G}$ as:
\[
\text{LICOM}(g; \tau) := \max_{\mathbf{x}_0(t_d ) \in g} \licp{t_d}{\mathbf{x}_0(t_d)}
\]

\end{Def}

\textbf{Practical Computation with Thresholding.}  
From $(\lambda,\tau)$-safety in ~Def.~\ref{safety}, we have a threshold parameter \( \lambda \in [0, 1] \) to prune low-probability regions during the maximization. Specifically, for each cell \( g \in \mathcal{G} \), we only consider ego poses \( \mathbf{x}_0(t_d) \in g \) that lie within regions where the obstacle's predicted probability density \( \mathds{P}(\mathbf{x}_i(t_d) \mid \mathbf{x}_i(t)) \) exceeds \( \lambda \).
This thresholding acts as a soft truncation of the obstacle's motion distribution, significantly reducing integration cost by ignoring low-likelihood obstacle states that have negligible contribution to the final risk estimate. 

\section{Closed-Loop Simulation Evaluation}

To jointly evaluate the spatial correctness and temporal feasibility of decisions under latency, we integrate \algname with CARLA to perform closed-loop simulation in dynamic traffic scenarios. To simulate latency, \algname runs CARLA in synchronous mode. The AV continues to follow its default planner and controller for a duration equal to the measured latency before applying the delayed operator decision. This setup enables repeatable and deterministic evaluation of how delayed human input affects safety, collision risk, and task completion across realistic driving scenarios.

\subsection{Experimental Setup}

We evaluate \algname\  using three traffic conflict scenarios, corresponding to the most common intersection cases among the 14 classes defined by Parker and Zegee \cite{parker1989traffic_conflict}.  
The scenarios are implemented in the Town03 map in CARLA. 
In all cases, motion prediction for dynamic obstacles is performed using an Extended Kalman Filter (EKF). The ego vehicle follows a fixed reference trajectory consisting of a straight segment, a maneuver (merge or turn), and a post-maneuver straight segment.
The simulator runs at 100 frames per second with a planning horizon of 40 steps, 0.01 seconds apart.
For each configuration, we run 100 independent trials in CARLA. Each trial samples randomly initial positions and velocities for ego-vehicle and obstacle vehicles.  The reference trajectory of the ego-vehicle is collected by running the full simulation without the obstacle vehicle. We consider the following scenarios:

\textit{Scenario I: Should I Merge?} 
The ego vehicle merges into a lane with oncoming traffic, starting from a position 90~m before the merge point. The maneuver consists of 5~s of straight driving, 2.5~s of merging, and 2~s of post-merge straight driving. 

\textit{Scenario II: Should I Turn Right?}
The ego vehicle turns to a right lane at an intersection, starting 90~m before the turn. An obstacle vehicle approaches from the top of the intersection, 30~m away. The ego follows a trajectory of 5~s straight driving, 2.5~s turning, and 2~s of post-turn motion.

\textit{Scenario III: Should I Turn Left?}
The ego vehicle turns to a left lane at an intersection, also starting 90~m before the turn, while an obstacle vehicle approaches from the opposite direction. The motion profile mirrors Scenario II, with 5~s of straight driving, 2.5~s turning, and 2~s post-turn.

We simulate best-effort human decision by computing the perceived collision probability between vehicles based on the state presented by \algname. 
If the perceived collision probability is more than the threshold for $\lambda$ as 0.3, and the human operator withhold the turning action. The local planner of the vehicles executes a braking control immediately. 

We consider a \textit{Baseline} method that does not account for human decision latency or compensate for its effects. In this setting, the simulated human operator responds to each visual query based solely on the current scene state, assuming immediate execution of the decision.

\begin{figure}
    \centering
    \begin{subfigure}[b]{0.32\linewidth}
        \centering
        \includegraphics[width=\linewidth]{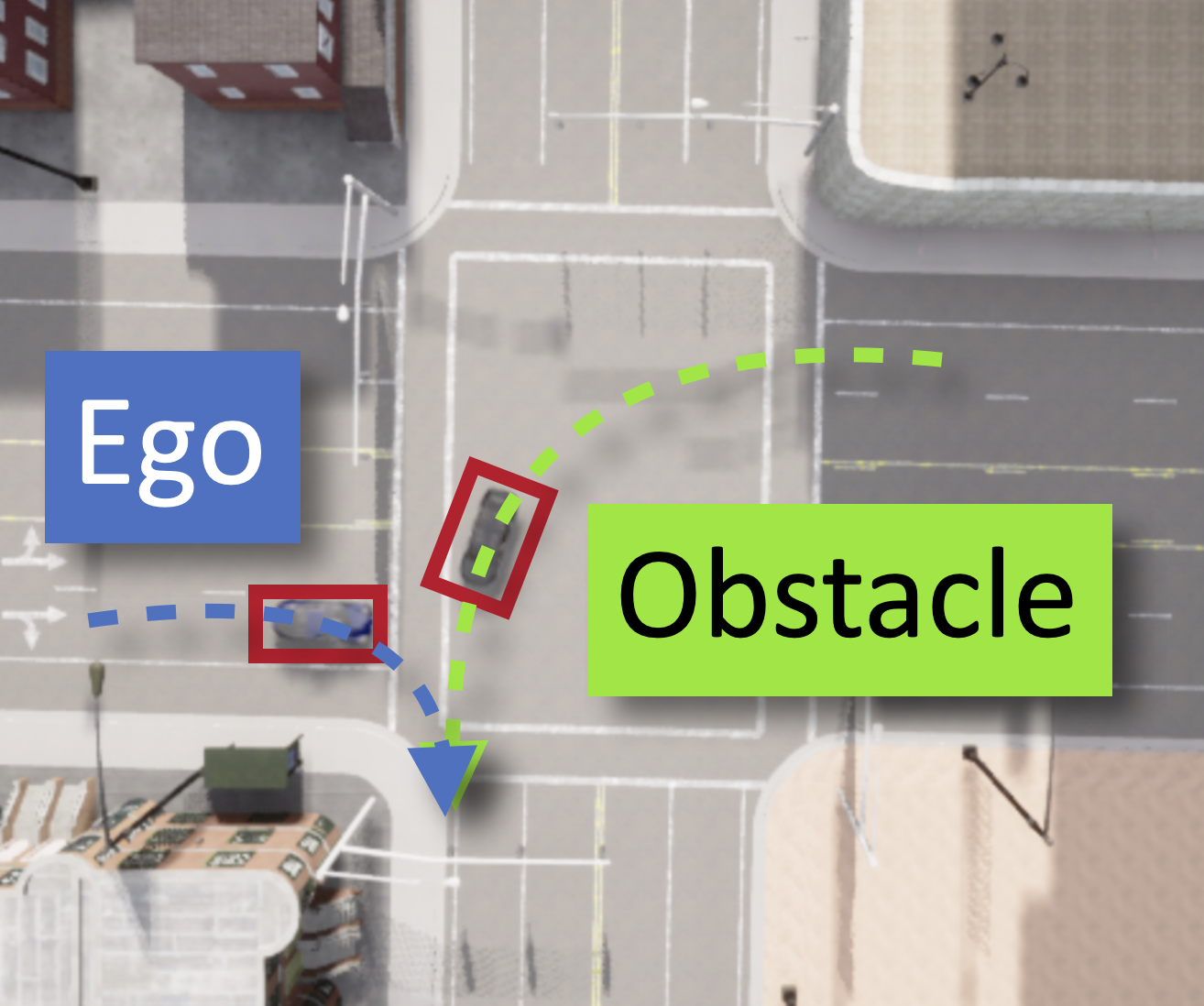}
        \caption{Should I merge?}
        \label{fig:setup-merge}
    \end{subfigure}
    \hfill
    \begin{subfigure}[b]{0.32\linewidth}
        \centering
        \includegraphics[width=\linewidth]{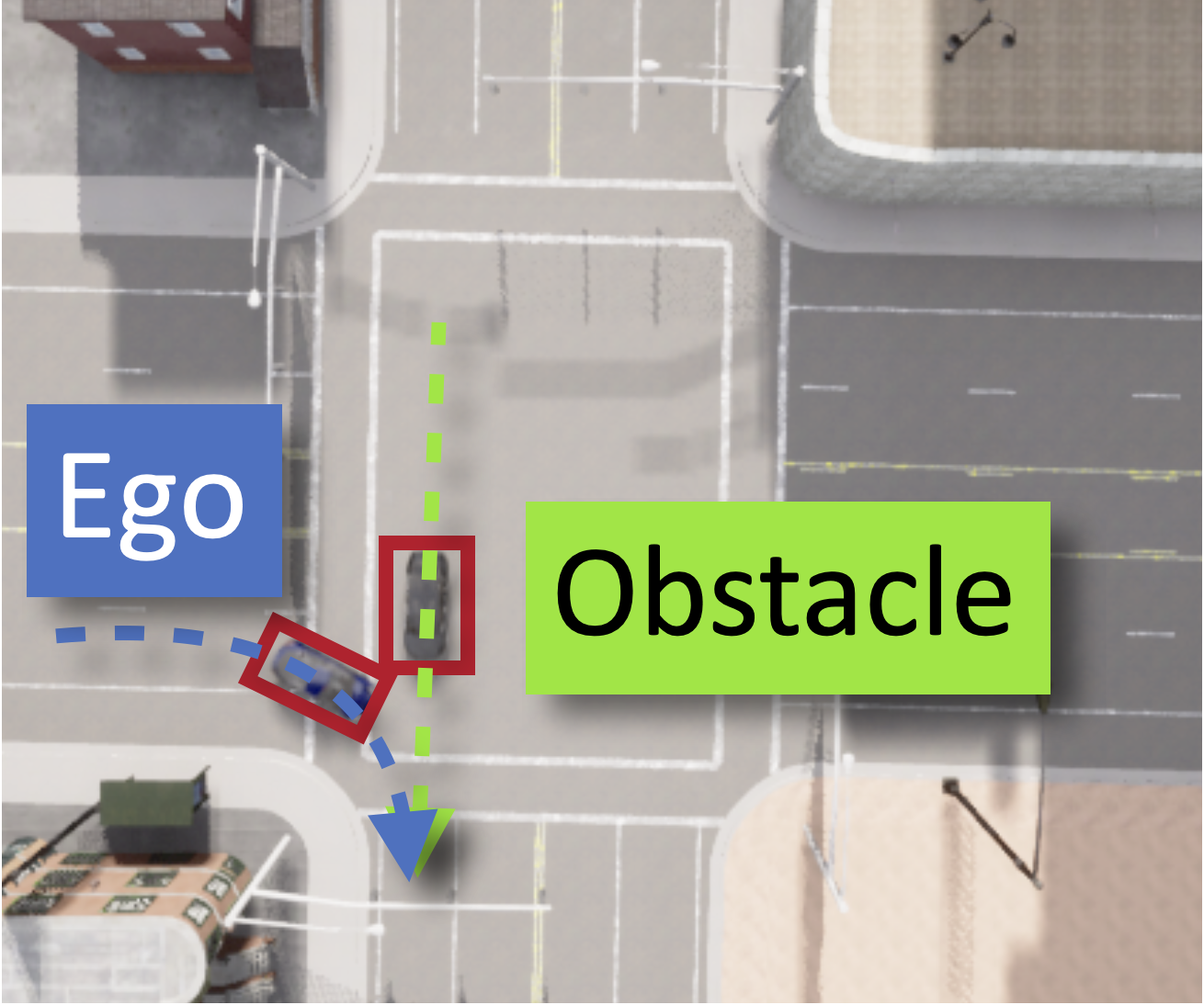}
        \caption{Should I right turn?}
        \label{fig:setup-right-turn}
    \end{subfigure}
    \hfill
    \begin{subfigure}[b]{0.32\linewidth}
        \centering
        \includegraphics[width=\linewidth]{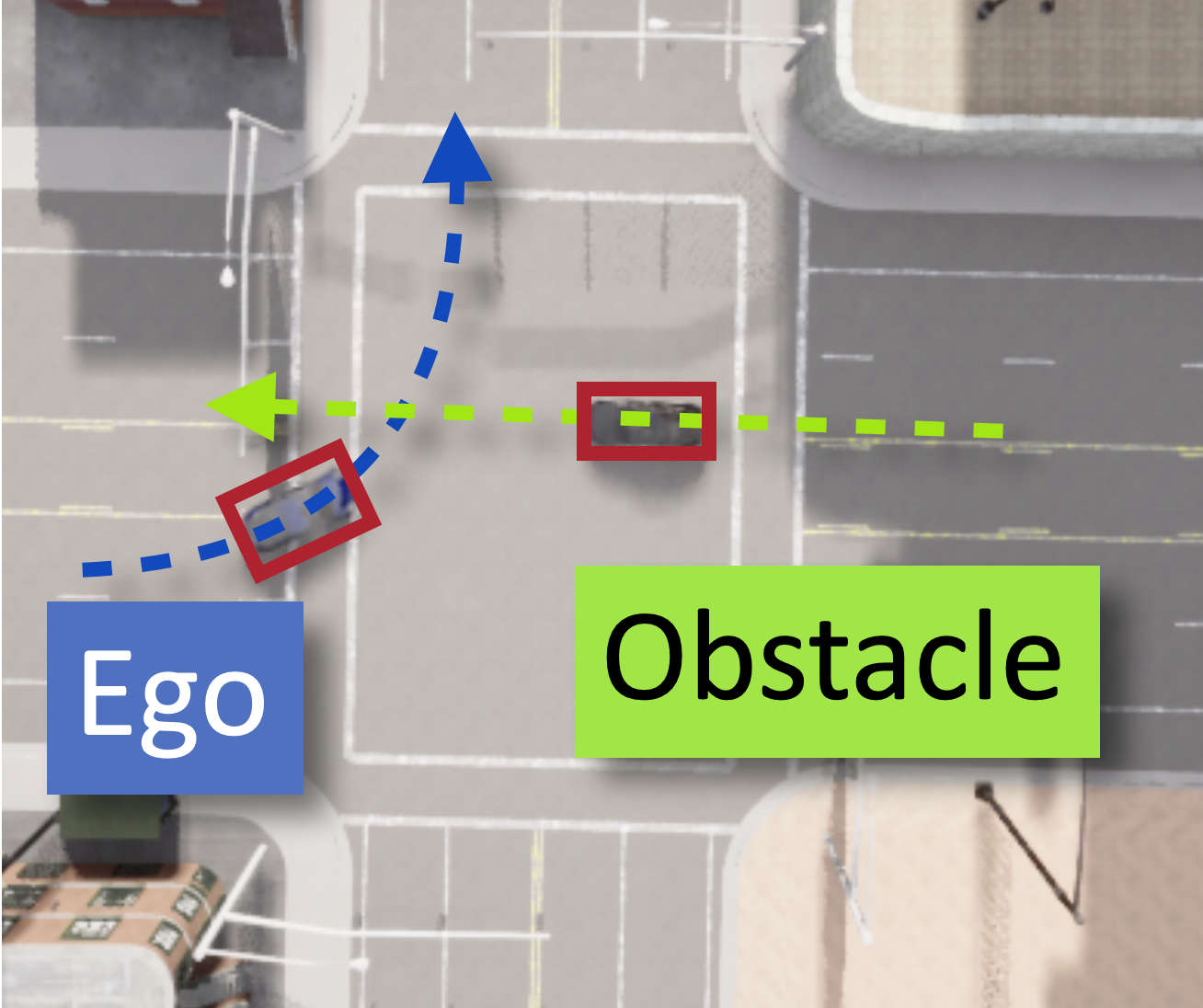}
        \caption{Should I left turn?}
        \label{fig:setup-left-turn}
    \end{subfigure}
    \caption{\textbf{Simulation Setup for Three Traffic Scenarios} that the AV must navigate in the presence of a dynamic obstacle.  }
    \label{fig:simulation-setups}
\end{figure}

\subsection{Collision Analysis}

\textit{Scenario I: Should I Merge?}
Table~\ref{tab:collision-rates-merge}  summarizes the collision rates for Baseline VQA that does not consider LICR and \algname methods with different latencies. In this merging scenario, the Baseline VQA without considering latency leads to collisions (100\% rate at 200--400ms). By contrast, \algname substantially reduces collisions at lower latency (16\% at 200ms).

\begin{table}
\centering
\footnotesize
\begin{tabular}{@{}lcc r@{}}
\toprule
Latency (ms) & Method & Collision Rate & Reduction Ratio \\
\midrule
    \multirow{2}{*}{200} 
      & Baseline     & 100\%    & 1x\\
      & \algname      & \textbf{16.0\% }  & \textbf{6.25×} \\ \addlinespace[.2em]
    \multirow{2}{*}{300} 
      & Baseline     & 100\%    &1x \\
      & \algname      & \textbf{37.0\%}   &  \textbf{2.70× }\\ \addlinespace[.2em]
    \multirow{2}{*}{400} 
      & Baseline     & 100\%    &  1x \\
      & \algname      & \textbf{66.0\%}   & \textbf{1.52× }\\
\bottomrule
\end{tabular}
\caption{{Collision Rates for Opposite Direction Merge}}
\label{tab:collision-rates-merge}
\end{table}

\textit{Scenario II: Should I Turn Right?}
Table~\ref{tab:collision-rates-right} 
 summarizes the collision outcomes under different latency levels for both the Baseline VQA and \algname methods. As latency increases from 200ms to 400ms, the Baseline approach shows higher collision rates, rising to 97.3\% at 400ms. The \algname method consistently lowers collision occurrences, especially at lower latency (e.g., 9.0\% at 200ms).

\begin{table}
\centering
\footnotesize
\begin{tabular}{@{}lcc r@{}}
\toprule
Latency (ms) & Method & Collision Rate & Reduction Ratio \\
\midrule
    \multirow{2}{*}{200} 
      & Baseline     & 80.8\%   &1x  \\
      & \algname      & \textbf{9.0\% }   &\textbf{ 8.98x}  \\ \addlinespace[.2em]
    \multirow{2}{*}{300} 
      & Baseline     & 82.8\%   &1x  \\ 
      & \algname      & \textbf{34.0\% }  & \textbf{2.41× }\\ \addlinespace[.2em] 
    \multirow{2}{*}{400} 
      & Baseline     & 97.3\%   &1x  \\
      & \algname      & \textbf{72.8\% }  & \textbf{1.34×} \\
\bottomrule
\end{tabular}
\caption{{Collision Rates for Right Turn} }
\label{tab:collision-rates-right}
\end{table}

\textit{Scenario III: Should I Turn Left?}
Table~\ref{tab:collision-rates-left} presents collision rates under increasing latency for both the Baseline VQA and \algname methods. At 200ms latency, the Baseline VQA approach incurs a 49.0\% collision rate, whereas \algname reduces it to 12.0\% (a 4.08× reduction).

\begin{table}
\centering
\footnotesize
\begin{tabular}{@{}lcc r@{}}
\toprule
 Latency (ms) & Method & Collision Rate & Reduction Ratio \\
\midrule
    \multirow{2}{*}{200} 
      & Baseline     & 46.0\%   & 1× \\
      & \algname      & \textbf{12.0\%}   & \textbf{4.08× }\\ \addlinespace[.2em]
    \multirow{2}{*}{300} 
      & Baseline     & 49.0\%   & 1×\\
      & \algname      &\textbf{ 16.3\% }  &\textbf{ 2.82× } \\ \addlinespace[.2em]
    \multirow{2}{*}{400} 
      & Baseline     & 53.8\%   &  1×\\
      & \algname      & \textbf{37.1\%}   &\textbf{ 1.45×} \\
\bottomrule
\end{tabular}
\caption{{Collision Rates for Left Turn} }
\label{tab:collision-rates-left}
\end{table}

\begin{figure*}[t]
    \centering
    \begin{subfigure}[b]{0.3\linewidth}
        \includegraphics[width=\linewidth]{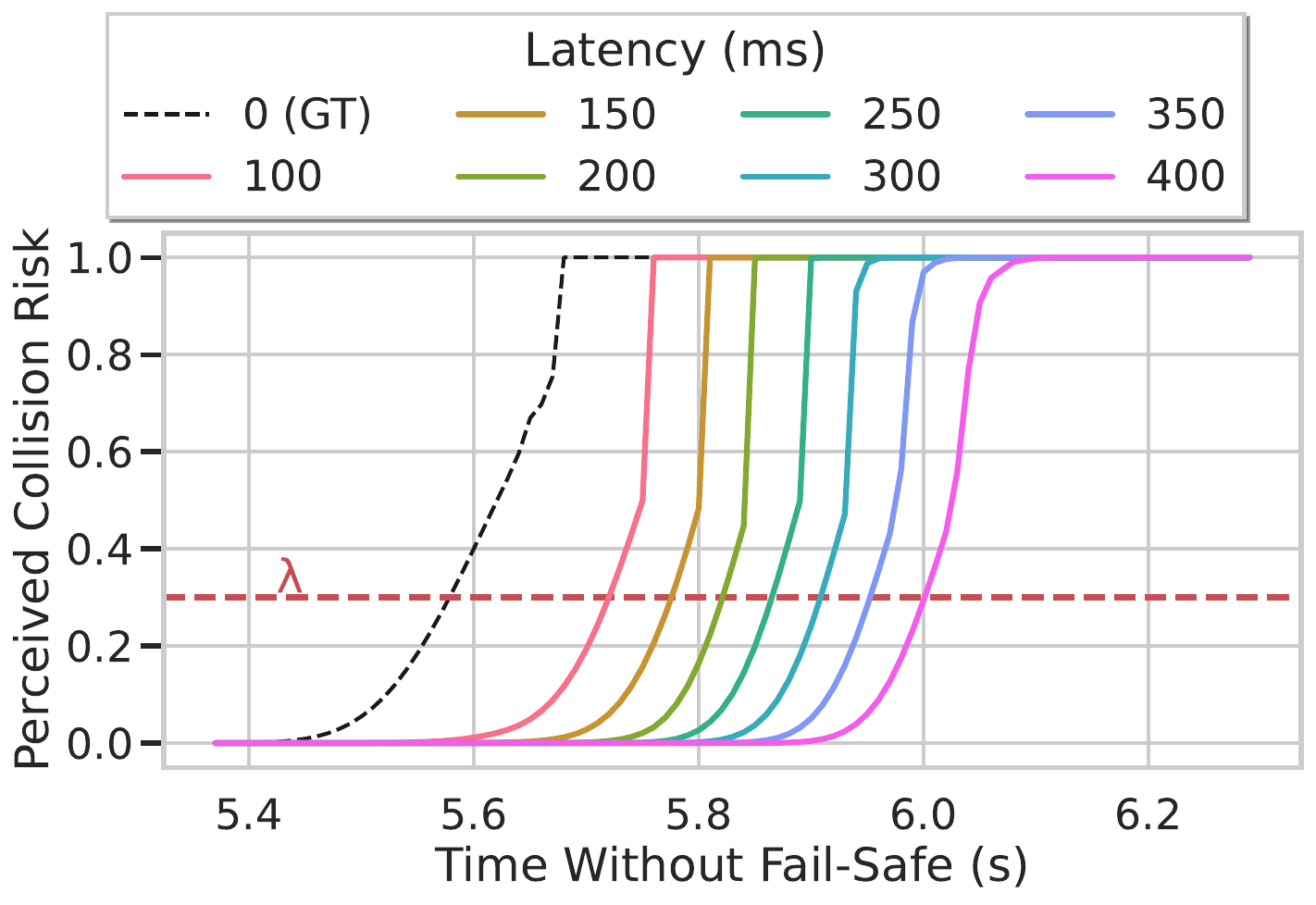}
        \caption{Merge}
        \label{fig:collision-merge}
    \end{subfigure}
    \hfill
    \begin{subfigure}[b]{0.3\linewidth}
        \includegraphics[width=\linewidth]{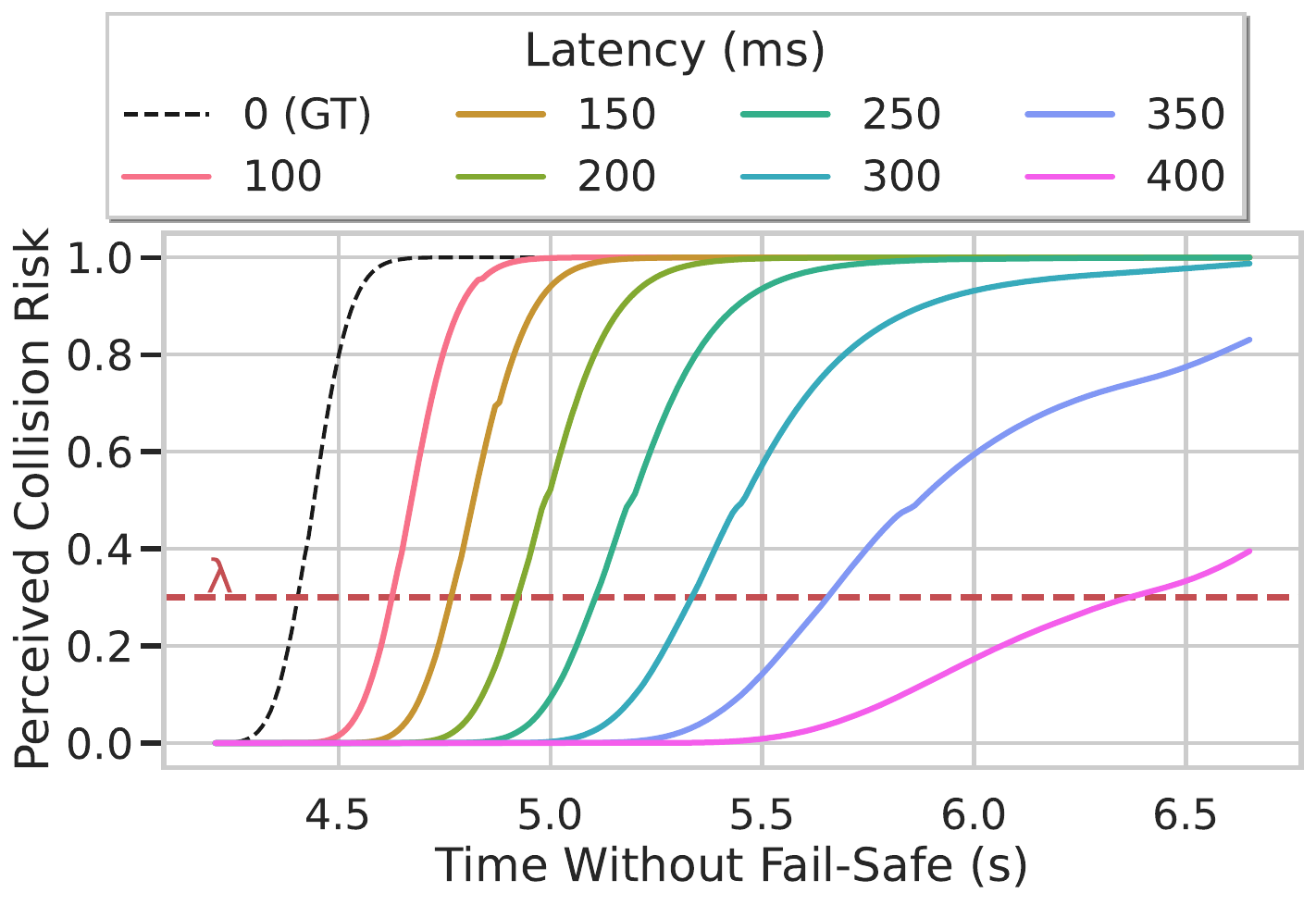}
        \caption{Right Turn}
        \label{fig:collision-probabilities-right-turn}
    \end{subfigure}
    \hfill
    \begin{subfigure}[b]{0.3\linewidth}
        \includegraphics[width=\linewidth]{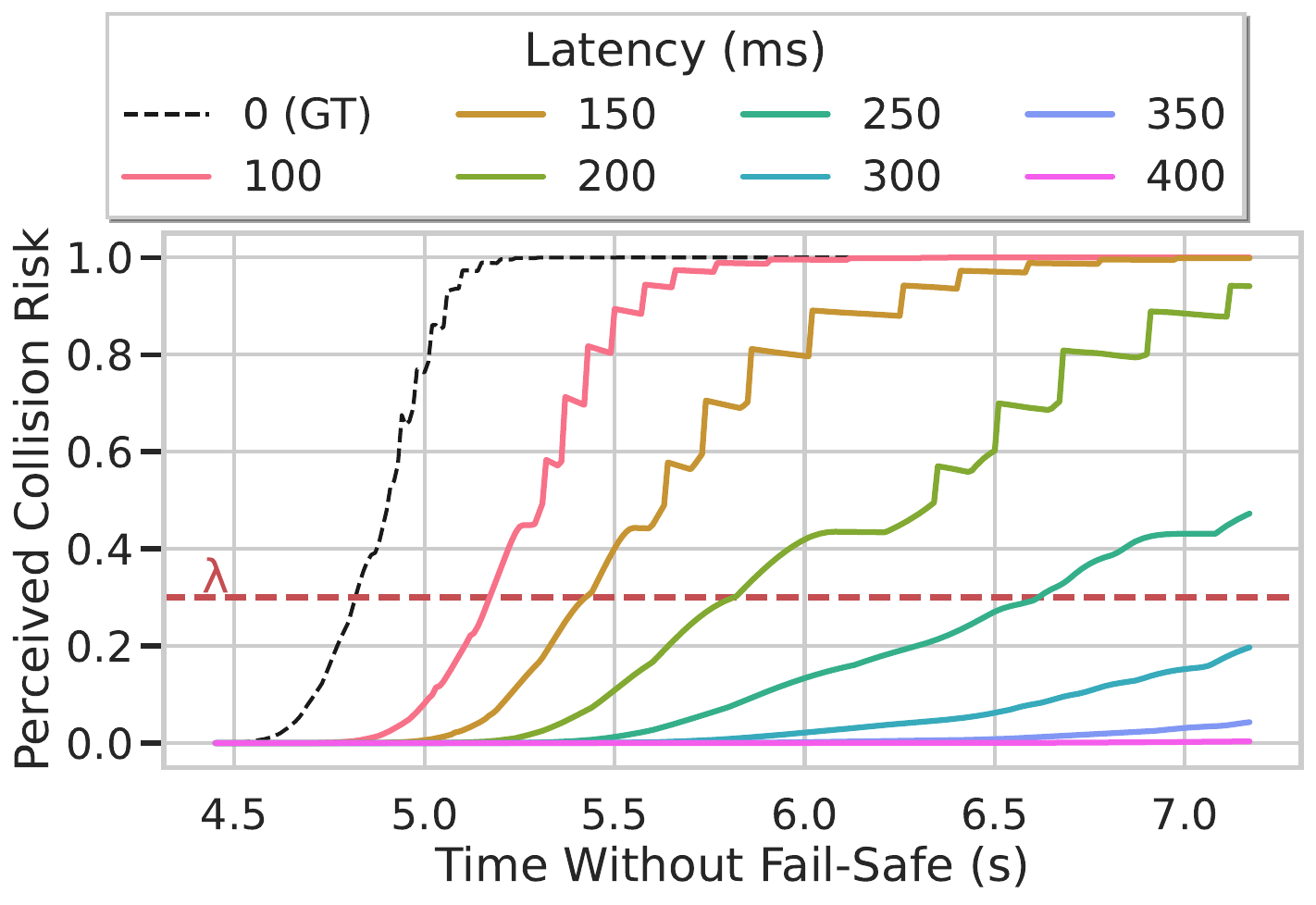}
        \caption{Left Turn}
        \label{fig:collision-probabilities-left-turn}
    \end{subfigure}
    \caption{\textbf{Perceived Collision Probability Across Scenarios and Latencies.} LICP models how 
    human perceives collision risk over time under varying latencies (0–400\,ms) in three traffic scenarios. Without visualizing and compensating latency-induced collision probability, higher latencies cause increasingly delayed perception of risk. Dashed horizontal lines indicate the risk threshold $\lambda$ that human make the decision.}
    \label{fig:collision-all}
\end{figure*}

\subsection{Perceived Collision Probability  Over Different Latencies}

We analyze the difference between the Baseline approach and \algname. 
Under different latencies, we compare 
the perceived LICP under different human decision delays. Across all three scenarios in Figure~\ref{fig:collision-all}, higher latencies consistently lead to delayed detection of risk and, in some cases, failure to identify critical danger before collision.

\textit{Scenario I: Should I Merge?}:  
In Figure~\ref{fig:collision-merge}, ground-truth collision risk rises sharply starting at 5.4\,s and peaks near 5.7\,s. With 100\,ms latency, the system detects risk with a delay of 0.12\,s and registers high risk ($\text{LICP} > \lambda$) after an additional 0.08\,s. At 400\,ms, delays increase to 0.48\,s for initial detection and 0.40\,s for high-risk identification. These curves shift rightward with latency, illustrating slower response to imminent collision.

\begin{figure*}
    \centering
    \includegraphics[width=\linewidth]{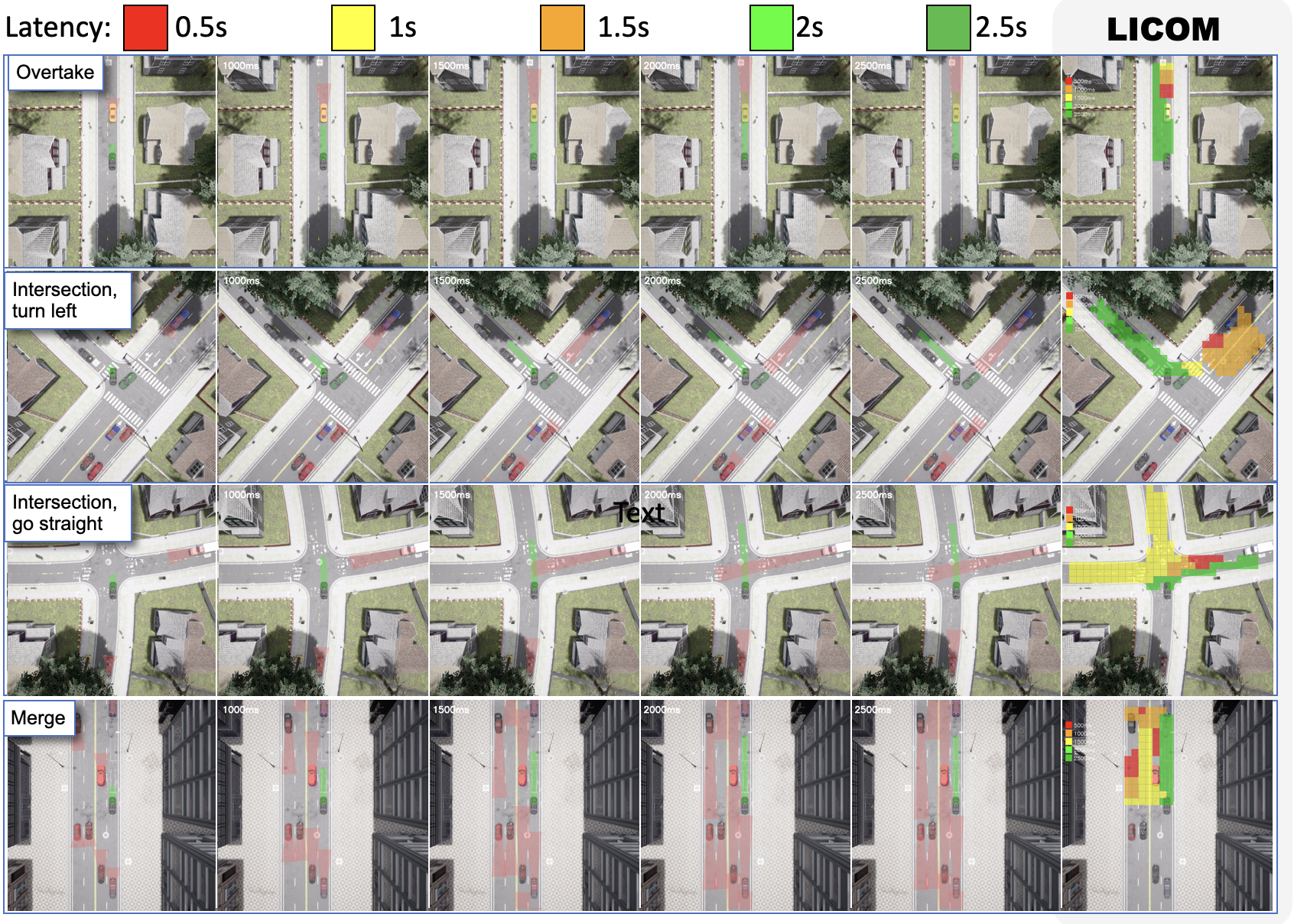}
    \caption{\textbf{Latency-Induced Collision Maps (LICOM) Across Driving Scenarios and Latency Levels.} Each row illustrates a different driving scenario in CARLA: \textit{Overtake}, \textit{Intersection – Turn Left}, \textit{Intersection – Go Straight}, and \textit{Merge}. Columns correspond to increasing decision latency from 0.5s to 2.5s. The final \mapname overlays indicate regions of high collision probability under delayed execution, with red denoting unsafe zones and green denoting safe zones. As latency increases, safe regions shrink or shift, highlighting the importance of accounting for temporal risk in dynamic environments.}
    \label{fig:visualization}
\end{figure*}

\textit{Scenario II: Should I Turn Right?} 
As shown in Figure~\ref{fig:collision-probabilities-right-turn}, the LICP remains negligible during straight motion and surges at 4.5\,s when the right turn begins. At 100\,ms latency, delays of 0.18\,s (initial risk) and 0.29\,s (high risk) are observed. These grow to 0.41\,s and 0.75\,s at 200\,ms. At latencies of 300\,ms or greater, the system detects only the onset of risk but misses the threshold-crossing event before collision, demonstrating unsafe perception under delay.

\textit{Scenario III: Should I Turn Left?} 
In Figure~\ref{fig:collision-probabilities-left-turn}, ground-truth risk rises at 4.5\,s and saturates by 4.9\,s. At 100\,ms latency, detection delays are 0.22\,s (risk onset) and 0.71\,s (high risk). At 150\,ms, these widen to 0.36\,s and 1.58\,s. Beyond 200\,ms, the system still registers initial risk but consistently fails to perceive high risk before impact. We note a slight difference compared to earlier case studies is the zig-zag pattern of the perceived collision risk. This is because the left turn is more abrupt compared to other scenarios, which EKF may not predict reliably.

\section{LICOM Visualization}
In this section, we qualitatively evaluate the spatial-temporal evolution of risk using the proposed \mapname. The \mapname is visualized within the CARLA Leaderboard framework.
At each decision step, the \mapname is computed over a discretized bird's-eye-view grid, where each cell encodes the maximum collision probability for ego positions within that region, given a fixed response latency.

Figure~\ref{fig:visualization} shows example snapshots from CARLA scenarios with \mapname.  
The high-risk regions (in red) expand and shift in alignment with predicted obstacle motion. The shrinking safe zone clearly illustrates the urgency of decision-making under delay, supporting the hypothesis that visualizing latency-induced risk can guide more temporally-aware operator responses.



\section{Conclusion}

This paper introduces \algname, a novel  framework for teleoperating autonomous driving that jointly models spatial risk and decision latency. We propose Latency-Induced Collision Probability (LICP), a measure that quantifies both temporal latency and spatial uncertainty, and Latency-Induced COllision Map (\mapname), which visualizes  how vehicle safety regions vary over time in the
presence of dynamic obstacles and human delay. 
Our experimental results demonstrate that our approach can reduce collision rates up to a 8.98x compared to a baseline that doesn not consider latency. 
We consider it as future work to integrate Vision-Language Models (VLM), an emerging technology for automating VQA, to the proposed framework.

\renewcommand*{\bibfont}{\small}
\printbibliography
\end{document}